\documentclass[journal]{IEEEtran}
\usepackage{epsfig}
\usepackage{graphicx}
\usepackage{amsmath}
\usepackage{amssymb}
\usepackage{url}
\usepackage{verbatim}
\usepackage{color}

\usepackage[breaklinks=true,colorlinks,bookmarks=false]{hyperref}

\ifCLASSINFOpdf
\else
\fi
\begin{document}
%
\title{Shape Prior Non-Uniform Sampling Guided Real-time Stereo 3D Object Detection}
%
%
%

\author{Aqi~ Gao, Jiale~Cao, and~Yanwei~Pang

\thanks{A. Gao, J. Cao, and Y. Pang are with the Tianjin Key Laboratory of Brain-Inspired Intelligence Technology, School of Electrical and Information Engineering, Tianjin University, Tianjin 300072, China (e-mail: (gaoaqi,connor,pyw)@tju.edu.cn).
}
}

\maketitle

\begin{abstract}

Pseudo-LiDAR based 3D object detectors have gained popularity due to their high accuracy. However, these methods need dense depth supervision and  suffer from inferior speed. To solve these two issues, a recently introduced RTS3D builds an efficient 4D Feature-Consistency Embedding (FCE) space for 
the intermediate representation of object without depth supervision.
FCE space splits the entire object region into 3D uniform grid latent space for feature sampling point generation, which ignores the importance of different object regions. However, we argue that, compared with the inner region, the outer region plays a more important role for accurate 3D detection. To encode more information from the outer region, we propose a shape prior non-uniform sampling strategy that performs dense sampling in outer region and sparse sampling in inner region. As a result, more points are sampled from the outer region and more useful features are extracted for 3D detection. Further, to enhance the feature discrimination of each sampling point, we propose a high-level semantic enhanced FCE module to exploit more contextual information and suppress noise better. Experiments on the KITTI dataset are performed to show the effectiveness of the proposed method. Compared with the baseline RTS3D, our proposed method has 2.57\% improvement  on AP$_{3d}$ almost without extra network parameters. Moreover, our proposed method outperforms the state-of-the-art methods without extra supervision at a real-time speed.

\end{abstract}


\IEEEpeerreviewmaketitle

\section{Introduction}

\begin{figure}[t]
\centering

\includegraphics[width=1\linewidth]{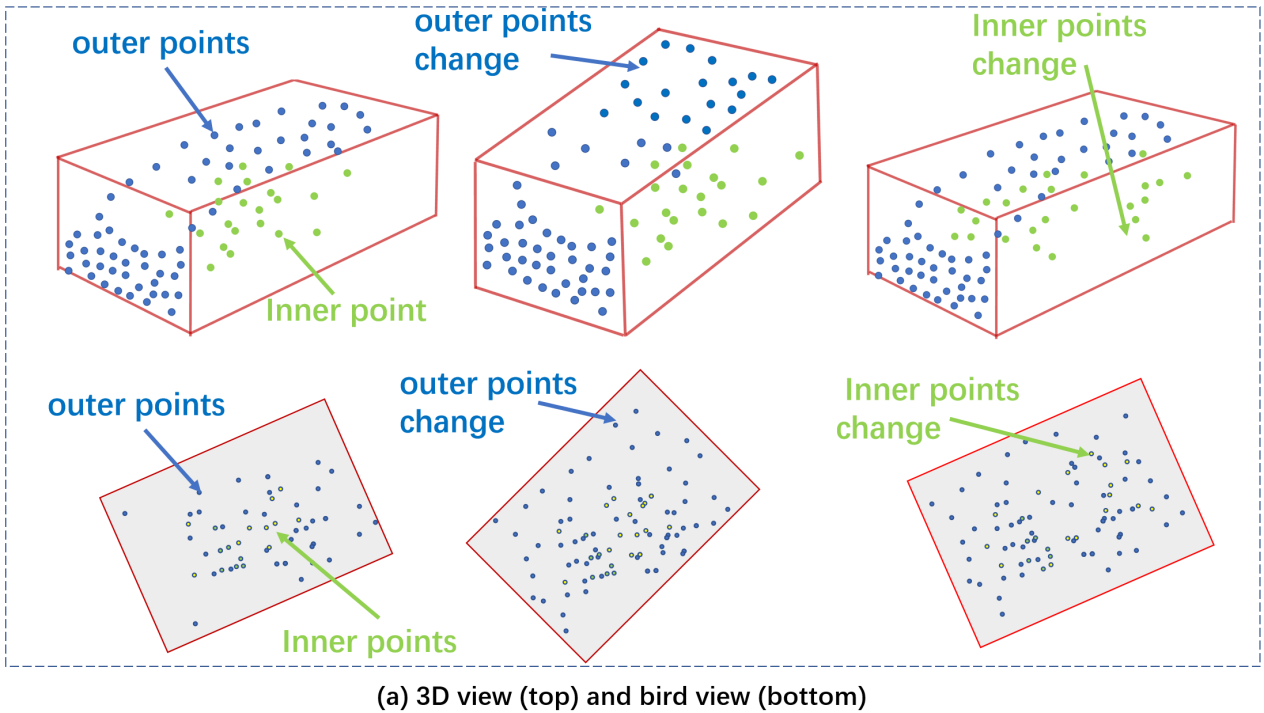}
\includegraphics[width=1\linewidth]{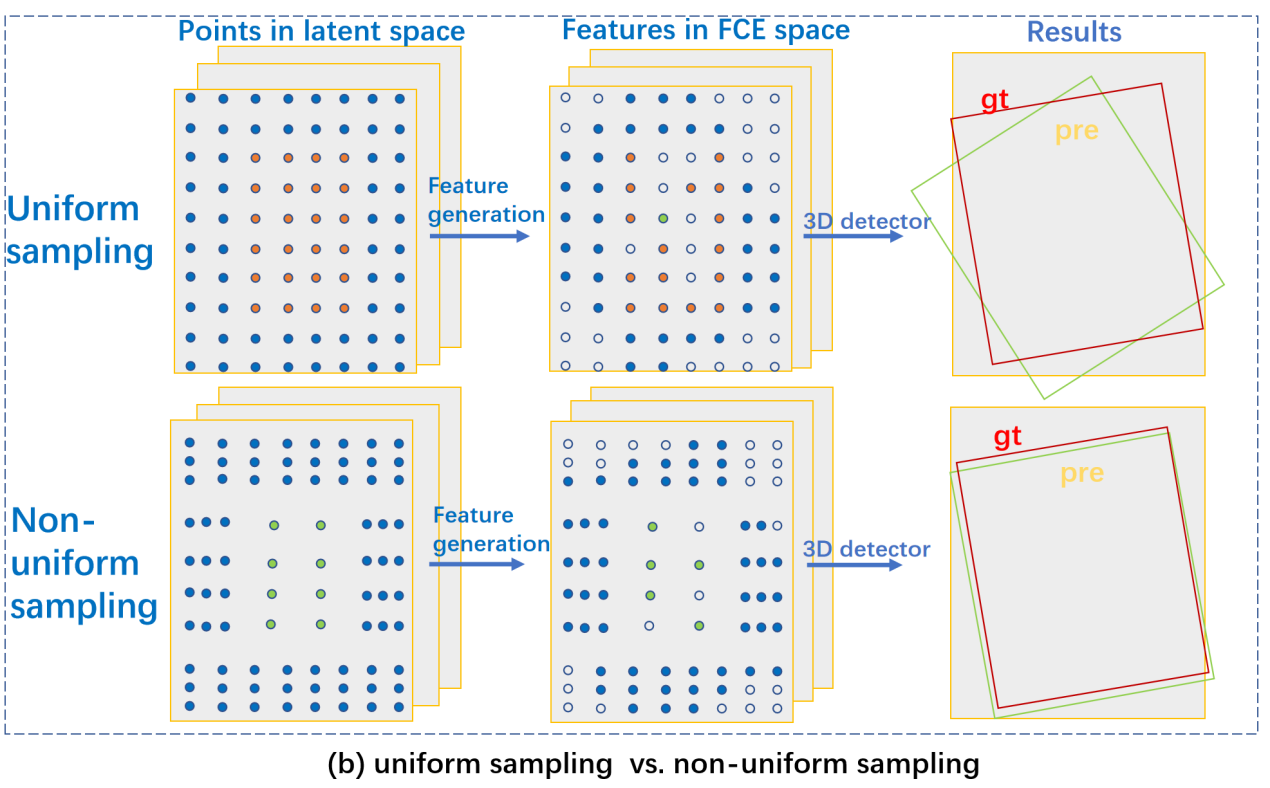}
\caption{In (a), we provide a  conceptual illustration of the importance of object (car) outer region for 3D detection. The first column shows the 3D points of a car in 3D view (top) and bird view (bottom).  When the outer points of the car change (second column), the corresponding 3D/2D bounding boxes (Ground-Truth) also change in both shape and orientation. However, when the inner points change (third column), the corresponding 3D/2D bounding boxes do not change. It demonstrates that the outer points of object play a more role for 3D detection. In (b), we compare uniform sampling and non-uniform sampling for FCE space generation. Compared to uniform sampling, our non-uniform sampling can generate more sampling points in outer region (first column). In the following FCE space, these outer points (second column) provide more useful features for improving the following 3D detection (third column).}
\label{fig:OuterExamples}
\end{figure}

\IEEEPARstart{3}{D} object detection is an important and fundamental task for automatic driving. The related methods can be mainly divided into LiDAR-based 3D object detection approaches \cite{Qi_PointNet_CVPR_2017,Qi_PointNet++_NIPS_2017,2019PointRCNN,2021pcgraph} and image-based 3D object detection approaches \cite{Li_StereoRCNN_CVPR_2019,Peng_IDA3D_CVPR_2020,wang_PseudoLiDAR_2019}.  Though LiDAR-based 3D object detection approaches have high accuracy, they suffer from the expensive hardaware cost and 
are sensitive to severe weather (\textit{e.g.,} rain and snow).  Compared with LiDAR-based 3D object detection approaches, image-based 3D object detection approaches adopt the low-cost optical camera and can provide dense depth information. Image-based 3D objection can be further divided into monocular 3D object detection and stereo 3D object detection. In this paper, we focus on real-time stereo 3D object detection.

Stereo 3D object detection is aimed at predicting 3D bounding boxes of objects using the stereo pairs of images.
With the technique of deep convolutional neural networks \cite{Krizhevsky_ImageNet_NIPS_2012,Simonyan_VGG_arXiv_2014,He_ResNet_CVPR_2016}, stereo 3D object detection has achieved great success in recent few years. Among the stereo 3D object detection methods, Pseudo-LiDAR based 3D object detection approaches\cite{wang_PseudoLiDAR_2019,you_Pseudo-LiDAR++_2020,xu_Zoomnet_2020,sun_disprcnn_2020}  are one of the most representative classes. Generally, Pseudo-LiDAR based approaches first predict the disparity map, second transform the disparity map into point cloud, and third employ a point cloud detector for 3D detection. Despite of high accuracy, these methods require pixel-wise depth labels and have a relatively slow inference speed, which limits the application in automatic driving. To solve these mentioned drawbacks in Pseudo-LiDAR based approaches, a recently proposed RTS3D \cite{Li_RTS3D_AAAI_2021} builds a 4D feature-consistency embedding (FCE) space as the intermediate representation of object. Specifically, RTS3D adopts uniform sampling to generate feature sampling points (grid) for each proposal and represents each point (grid) with consistency features generated from stereo images. Based on the features generated in FCE space, a 3D detector is employed for 3D bounding prediction. With these simple and efficient designs, RTS3D not only avoids the pixel-wise depth supervision, but also achieves a very competitive accuracy at a real-time speed. 

Despite the success, we argue that there are some inappropriate designs in RTS3D that impede its performance. First, RTS3D adopts uniform sampling strategy to generate feature sampling points, which ignores the importance of different object regions. As shown in Fig. \ref{fig:OuterExamples}(a), compared to the inner points from the inner region, the outer points from the outer region play a more important role for 3D detection. However, the uniform sampling strategy adopted in RTS3D pays an equal attention to different object regions.  Second, RTS3D does not fully exploit the contextual information to suppress noise during the consistency feature generation.

To address the issues in the state-of-the-art detector RTS3D, we propose a Shape Prior non-uniform Sampling guided 3D detector, called SPS3D. Instead of constructing a uniform 3D grid space for each 3D proposal, we propose  to build a non-uniform 3D grid latent space by considering object shape prior information. In each dimension of the 3D proposal, we design a piece-wise linear function to sample more points (grid) from outer region and less points from the inner region. After that, we extract the corresponding consistency feature for each sampling point to construct the non-uniform FCE space. As shown in Fig. \ref{fig:OuterExamples}(b), our proposed non-uniform sampling strategy exploit more useful features for the following 3D detection because  more sampling points from the outer region are extracted. In addition, to enhance feature discrimination for each point, we propose a high-level semantic enhanced FCE module that exploits more contextual information for feature representation. Finally, we adopt a 3D detector for 3D bounding box prediction. Overall, the contributions and merits of this paper are summarized as follows. 

\begin{itemize}
   
    \item We observe that outer region of an object plays a more important role for 3D bounding box prediction.  Further we propose the model that devides the outer region and inter region of the car.
    \item 
    Based on this model, we propose a shape prior non-uniform sampling mechanism for feature sampling point generation. As a result, more useful points from discriminative region can be sampled  for 3D bounding box prediction.
    \item To enhance the feature of each sampling point, we further introduce a high-level semantic enhanced FCE module to integrate more contextual information and suppress noise better.
    \item We validate the effectiveness and superiority of our proposed SPS3D on the challenging KITTI dataset \cite{kitti}. On the KITTI validation moderate set, our SPS3D outperforms the baseline RTS3D by an absolute gain of 2.57\% $AP_{3d}$ almost without additional computational costs. Moreover, our SPS3D achieves the state-of-the-art accuracy at real-time speed.
\end{itemize}

\section{Related work}

\subsection{2D object detection}
In past few years, deep convoultional neural networks have made great progress in 2D object detection \cite{Girshick_RCNN_CVPR_2014,Ren_FasterRCNN_NIPS_2015,Liu_SSD_ECCV_2016,Redmon_YOLO_CVPR_2016}. The object detection methods mainly consist of two-stage approaches \cite{Girshick_RCNN_CVPR_2014,Lin_FPN_CVPR_2017,Cao_D2Det_CVPR_2020,Li_TridentNet_ICCV_2019,Fang_cfr} and one-stage approaches \cite{Liu_SSD_ECCV_2016,Cao_HSD_ICCV_2019,Zhang_RefineDet_CVPR_2018,Liu_RFB_ECCV_2018,Lin_Focal_ICCV_2017}.  The two-stage approaches first extract some candidate class-agnostic proposals and second classify these proposals into specific classes, while the one-stage approaches directly predict class-ware bounding-boxes. At first, these object detection methods are anchor-based approaches with some handcrafted parameters. To avoid these handcrafted parameters, some anchor-free approaches are proposed recently, including key-point based approaches \cite{Law_CornerNet_ECCV_2018,Zhou_ExtremeNet_CVPR_2019,Duan_CenterNet_ICCV_2019} and center-point based approaches \cite{Tian_FCOS_ICCV_2019,Zhu_FSAF_CVPR_2019,Zhang_ATSS_CVPR_2020}.

\subsection{3D point cloud object detection}
3D point cloud object detection is crucial for automatic driving. To better perform 3D object detection, some deep backbones (\textit{e.g.,} PointNet \cite{Qi_PointNet_CVPR_2017} and PointNet++ \cite{Qi_PointNet++_NIPS_2017}) are proposed to extract the features from the point cloud.  Based on these bacbones, some 3D detectors (\textit{e.g.,} VoteNet \cite {Qi_voteNet_ICCV_2019} and MLCVNet \cite{Xie_MLCVNet_CVPR_2020}) are proposed.  VoteNet \cite {Qi_voteNet_ICCV_2019}
designs an end-to-end 3D object detection network based on a synergy of deep point set networks and Hough voting. 
MLCVNet \cite{Xie_MLCVNet_CVPR_2020} extracts multi-level contextual information with the self-attention mechanism and multi-scale feature fusion. Besides these methods, PointRCNN \cite{2019PointRCNN} constructs a two-stage detection framework for 3D detection. After that, many variants \cite{chen_FPR-CNN_ICCV_2019, 2019PointRGCN,Shi_PointGNN_CVPR_2020} are proposed. Recently, some works \cite{pan_Pointformer_2020, Guo_PointTransformer_2020,Zhao_PointTransformer_2020} apply transformer\cite{vaswani2017attention} to 3D point cloud detection.

\begin{figure*}
\begin{center}
\includegraphics[width=1\linewidth]{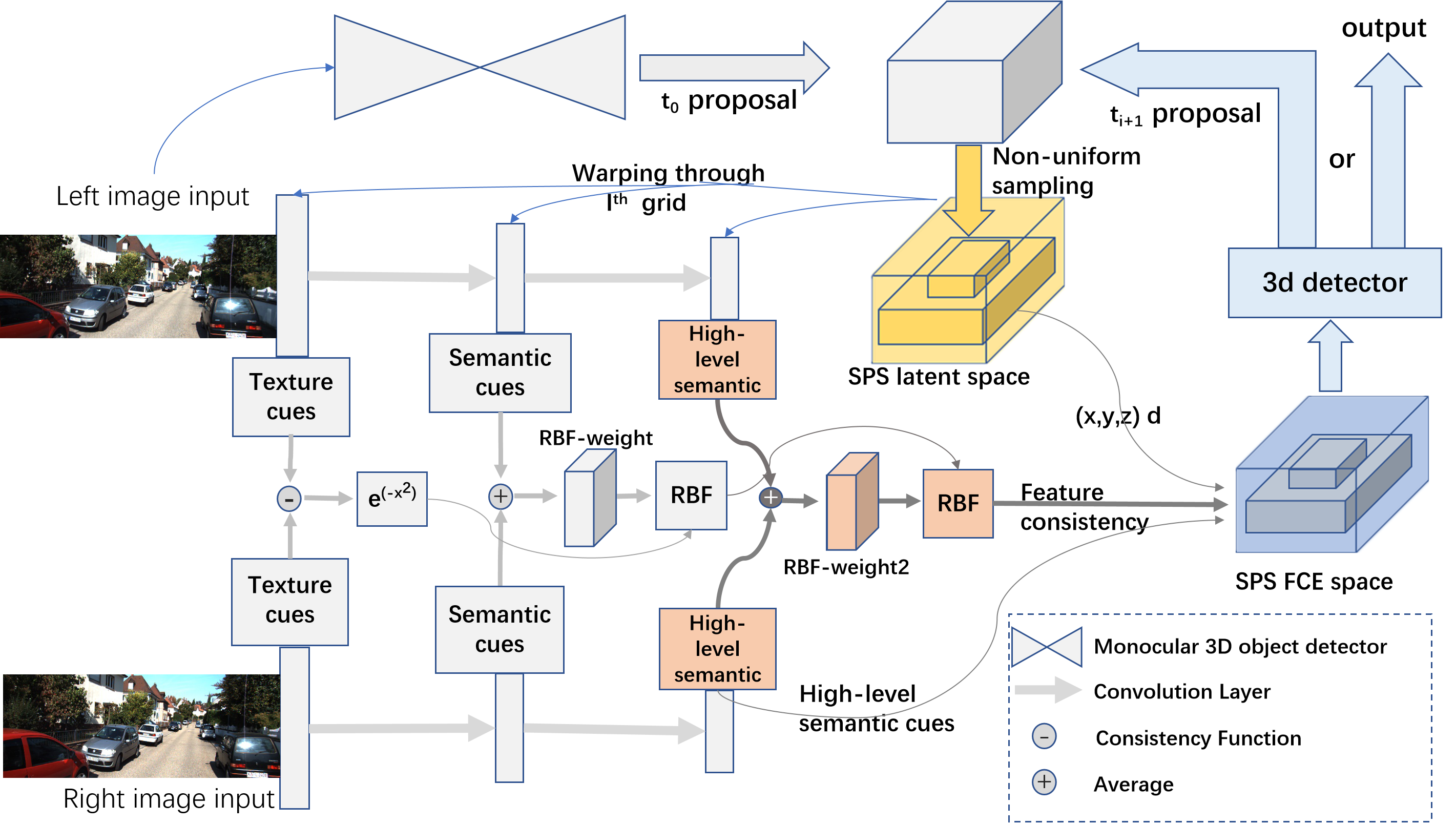}
\caption{Overall architecture of our proposed SPS3D for stereo 3D object (car) detection. First, we employ a fast  monocular  3D  detector to extract candidate 3D proposals. For each proposal, we perform shape prior non-uniform sampling to generate non-uniform 3D latent space, called SPS latent space. For each point in latent space, we extract the high-level enhanced consistency features to generate feature-consistency embedding (FCE) space and employ an improved 3D detector for 3D bounding box prediction. To improve detection performance, we perform multiple iterations, where the predicted 3D bounding box in current iteration is used as the input of next iteration.}
\label{fig:arch}
\end{center}
\end{figure*}

\subsection{Monocular 3D object detection}
Monocular 3D object detection aims to predict 3D bounding boxes from monocular image.  Mono3D \cite{Yan_Mono3D_TCS_2020,HE_Mono3D++_2017} first generates a 3D candidate box, second projects it to 2D scene, and third detects objects in 2D scene. 
Deep3Dbbox \cite{Mousavian_Deep3Dbbox_2017} proposes to use angle and scale information for depth estimation and 3D detection.
Deep MANTA \cite{Chabot_DeepMANTA_2017} defines a series of key points for a car and then uses the 3D template library for matching. KM3DNet \cite{Li_KM3D_ECCV} develops a novel single-shot and kepoints-based framework for monocular 3D objects detection. 
MonoFENet \cite{2019MonoFENet} estimates disparity from the
input monocular image, the features of both the 2D and 3D streams can be enhanced and utilized for accurate 3D localization. 
CaDDN \cite{Reading_CaDDN_cvpr} uses a predicted categorical depth distribution for each pixel to project rich contextual feature information to the appropriate depth interval in 3D space then get the final result. 
M3DSSD \cite{Luo_M3DSSD_CVPR} proposes a two-step feature alignment approach to overcome feature mismatching.  MonoRUN \cite{chen_MonoRUn_cvpr}  learns dense correspondences and geometry in a self-supervised manner with simple 3D bounding box annotations.

\subsection{Stereo 3D object detection}
Stereo 3D object detection mainly consists of two classes. Some methods need parallax and other supervision information. Pseudo-LIDAR \cite{wang_PseudoLiDAR_2019} is one of the representative methods. It transforms the depth map into point cloud and performs 3D point cloud detection. Pseudo-LIDAR++ \cite{you_Pseudo-LiDAR++_2020} proposes depth cost volume to get depth map directly. OC-Stereo \cite{Pon_ocstereo_2020} and Disp RCNN \cite{sun_disprcnn_2020} only consider point cloud coming from the foreground regions. ZoomNet \cite{xu_Zoomnet_2020} improves the effect of disparity estimation by enlarging the target. Some other methods do not need extra supervision. Stereo-RCNN  \cite{Li_StereoRCNN_CVPR_2019} generates a rough 3D bounding box by combining the RoIs from the left and right images and conducts BA optimization for final 3D bounding box prediction. IDA-3D \cite{Peng_IDA3D_CVPR_2020} builds cost volume from left and right ROI to get the depth of the center point for 3D detection. In this paper, we focus on stereo 3D object detection without using extra supervision information.

RTS3D \cite{Li_RTS3D_AAAI_2021} builds FCE space to represent the object.  Compared to Pseudo LIDAR \cite{wang_PseudoLiDAR_2019}, RTS3D achieves a better accuracy without dense supervision information. Moreover, it has a real-time speed. However, we argue that RTS3D ingores the importance of different regions for 3D detection.

\section{Method}

\begin{figure*}[t]
\begin{center}
\includegraphics[width=1\linewidth]{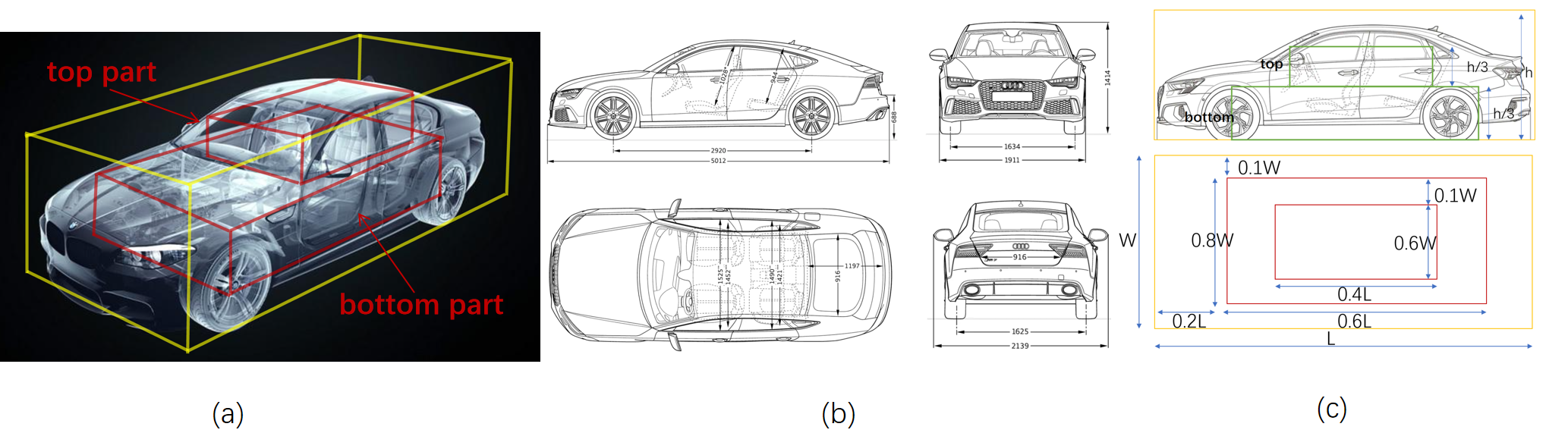}
\caption{(a) We show a 3D model of a car and corresponding groud-truth bounding-box (green). The car is divided into two cubes (red) for generating non-uniform sampling function. (b) The internal parameters of car model in four different views. (c) Detailed parameters (the width, length, height of two cubes) of two cubes (green). The top part shows the parameters in side view, and the bottom part shows the parameters in bird view.}
\label{fig:carmodel}
\end{center}
\end{figure*}

In this section, we provide a detailed introduction about our proposed method, called SPS3D, which is  built  on  real-time stereo 3D object detector RTS3D \cite{Li_RTS3D_AAAI_2021}. First of all, we give a review about RTS3D that consists of four steps: (1) 3D proposal generation. A efficient monocular 3D object detector is employed to extract some candidate 3D object proposals. (2) Multi-scale feature extraction for stereo images. The lightweight model ResNet-18 \cite{He_ResNet_CVPR_2016} is used to extract multi-scale feature maps. (3) Feature-Consistency Embedding (FCE) space generation for each proposal. RTS3D splits each 3D object proposal into uniform 3D grids and extracts the consistency features of each point (grid) from left and right multi-scale feature maps. As a result, each 3D proposal is represented by a 4D feature map. (4) 3D bounding box prediction. An improved PointNet \cite{Qi_PointNet_CVPR_2017} is designed for 3D bounding box prediction and confidence score estimation. The key step in RTS3D is  Feature-Consistency Embedding (FCE) space generation using uniform sampling strategy.

We argue that RTS3D has some inappropriate designs that impede the performance. The first one is uniform sampling for 3D grid space generation pays equal attention to all the object regions and thus ignores the importance of different regions. To pay more attention on the important regions, we propose a novel shape prior non-uniform sampling strategy. The second one is that consistency features for each sampling point are easily influenced by the noise. To better suppress the noise, we propose a high-level semantic enhanced FCE module to exploit more information. The overall architecture of our SPS3D is shown in Fig. \ref{fig:arch}. We first construct a shape prior non-uniform latent space, second generate non-uniform FCE space, and third perform 3D detection.

\subsection{Shape prior non-uniform latent space construction}
As discussed earlier, the different regions of objects have different importance. RTS3D adopts uniform sampling strategy to generate 3D latent space for each proposal. As a result, RTS3D ignores some points from the important (outer) region and generates many redundant points from the unimportant (inner) region. In fact, the points of the outer region play more important role in 3D bounding box prediction. For example, in Fig. \ref{fig:OuterExamples}(a), the 3D ground-truth bounding box of car is generated by the outer contour. Thus, we propose a shape prior non-uniform sampling strategy for latent space construction.

\begin{figure}[t]
\centering
\includegraphics[width=1\linewidth]{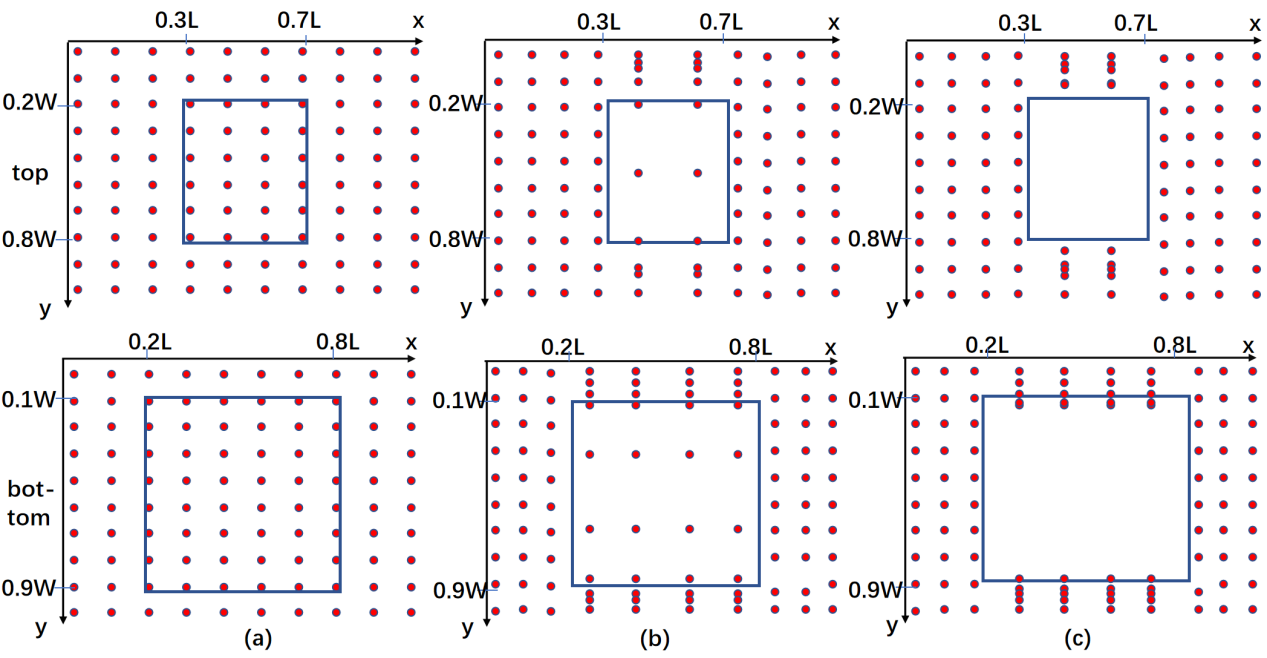}
\caption{Illustration of three different sampling methods (bird view). We show the sampling points in the top and bottom parts, divided by on two cubes, of the car. (a) is the uniform sampling method, (b) is our proposed non-uniform sampling method, and (c) is the extreme non-sampling method that does not generate sampling points in the inner regions.}
\label{fig:nonsampling}
\end{figure}

To simply perform shape prior non-uniform sampling, we need to model the shape of the car. 
We find that it is not necessary to use an accurate and unified mathematical formula to model the car. First, the proposal, generated by the monocular 3D detector, cannot give an accurate location of the car. Therefore, even an accurate model of the car cannot accurately distinguish the inner and outer parts inside the proposal. Second, although the appearances of different cars are different, some important parameter ratios (\textit{e.g.,} 
ratio between wheelbase and car length) are similar. Thus, we propose to use a single and simple model to represent all the cars in Fig. \ref{fig:carmodel}(a), where the car is divided into a top cube and a bottom cube. In this paper, we use the Audi car model\footnote{\url{https://www.audi.cn/cn/web/zh/models/a7/s7_sportback.html} }  to calculate the parameters(\textit{e.g.,} width, height, etc) of two cubes.  Fig.\ref{fig:carmodel}(b) shows some detailed shape parameters of an Audi car in four different views (\textit{i.e.,} side view, front view, bird view, back view). Based on these parameters in different views, we can generate two cubes. Fig. \ref{fig:carmodel}(c) shows the two cubes in bird view. The width and length of the car are represented as $W$ and $L$. Then, the width and length of top cube is 0.8$W$ and 0.6$L$, while the width and length of bottom cube is 0.6$W$ and 0.4$L$.

Based on these two cubes, we perform non-uniform sampling to build non uniform latent space for each 3D proposal. Fig. \ref{fig:nonsampling}(b) shows our non-uniform sampling in length ($x$-axis) and width ($y$-axis) directions. Here, we introduce how to generate the non-uniform sampling points for the bottom part of the car (see in the bottom part of Fig. \ref{fig:nonsampling}(b)).  When $x\leq0.2L$, the locations of sampling points can be written as
\begin{equation}
\left\{
\begin{array}{l}
X = ls(0, 0.2L, N_{x1}),\\
Y = ls(0, W, N_y),
\end{array}
\right.
\end{equation}
where $ls$ indicates the function of linespace, and $N_{x1}$ and $N_{y1}$ represnt the number of points in $x$-axis and $y$-axis directions. 

When $x>0.2L~\text{and}~x\leq 0.8L$, the $x$-axis locations of sampling points can be represented by $X = ls(0.2L, 0.8L, N_{x2})$, and the $y$-axis locations of sampling points can be written as
\begin{equation}
\left\{
\begin{array}{l}
Y = ls(0, 0.1W, N_{y1}),\\
Y = ls(0.1W, 0.9W, N_{y2}),\\
Y = ls(0.9W, W, N_{y3}).
\end{array}
\right.
\end{equation}

When $x>0.8L$, the locations of sampling points can be written as
\begin{equation}
\left\{
\begin{array}{l}
X = ls(0.8L, L, N_{x3}),\\
Y = ls(0, W, N_y).
\end{array}
\right.
\end{equation}

\begin{table}[t]
\begin{center}
\caption{The number of sampling points in $x$-axis and $y$-axis directions of three different sampling strategies.}
\begin{tabular}{l|c|c|c}
 \hline
Part& Method & $\{N_{x1},N_{x2},N_{x3}\}$ & $\{N_{y1},N_{y2},N_{y3}\}$ \\ \hline \hline
             & Fig. \ref{fig:nonsampling}(a)   & \{2,6,2\}   & \{1,8,1\}  \\ 
Bottom part  & Fig. \ref{fig:nonsampling}(b)    &  \{3,4,3\} &\{4,2,4\} \\     
             & Fig. \ref{fig:nonsampling}(c)   &  \{3,4,3\}  & \{5,0,5\}  \\ 
   
\hline
          & Fig. \ref{fig:nonsampling}(a) & \{3,4,3\} & \{2,6,2\} \\ 
Top part  & Fig. \ref{fig:nonsampling}(b) & \{4,2,4\} & \{4,3,3\}  \\     
          & Fig. \ref{fig:nonsampling}(c) & \{4,2,4\} & \{5,0,5\}  \\ \hline
\end{tabular}
\label{tab:numpoints}
\end{center}
\end{table}

In the similar way, we generate the non-uniform sampling points for the top part of the car (see in the top part of Fig. \ref{fig:nonsampling}(b)). We also show uniform sampling strategy in Fig. \ref{fig:nonsampling}(a) and the extreme non-sampling strategy in Fig. \ref{fig:nonsampling}(c). In extreme non-sampling strategy, we only generate the sampling points in the outer region. Compared to uniform sampling, our proposed non-sampling strategy pays more attention on the outer region. Compared to the extreme non-sampling strategy, our proposed non-sampling strategy does not ignore the inner region. Experimental results demonstrate that our non-sampling strategy is superior to both uniform sampling and extreme non-uniform sampling. It means that the outer region plays more important role than the inner region and the inner region is also useful for detection. Table \ref{tab:numpoints} further gives the number sampling points in both $x$-axis and $y$-axis directions for three different strategies, we set resl = 10.

With the proposed non-uniform strategy, we generate the non-uniform sampling points to construct the shape prior latent space for each 3D proposal. After that, we generate shape prior non-uniform FCE space for following 3D detection.

\begin{table*}[t]\scriptsize
\begin{center}
\caption{Comparison ($AP_{3D}$) of state-of-the-art 3D car detection methods on KITTI validation set. The number in the bracket indicate the improvement compared to RTS3D \protect\footnotemark
}
\resizebox{\linewidth}{!}{
\begin{tabular}{l|c|c|c|c|c|c|c|c}
\hline
Method & Extra supervision & Time &  \multicolumn{3}{c|}{IoU $ >$ 0.5} & \multicolumn{3}{c}{IoU $>$ 0.7}\\ \cline{4-9}
 &  &  & Easy & Moderate & Hard & Easy & Moderate &        Hard \\ \hline\hline
3DOP \cite{3DOP} & Instance Mask & - & 46 & 34.6 & 30.1 & 6.6 & 5.1 & 4.1 \\ 
MLF \cite{MLF} & Depth & - & - & 47.4 & - & - & 9.8 & - \\ 
YOLOstereo3d \cite{liu2021yolostereo3d}& Depth & 80ms & - & -& - &72.06 & 46.58 & 35.53\\
DSGN \cite{sun_disprcnn_2020}& Depth & 670ms & - & -& - &72.31& 54.27&47.71\\
PL: F-PointNet \cite{wang_PseudoLiDAR_2019} & Depth+Flow & 670ms & 89.5 &75.5 & 66.3 & 59.4 & 39.8 & 33.5 \\ 
PL: AVOD \cite{wang_PseudoLiDAR_2019} & Depth+Flow & 510ms & 88.5 & 76.4 & 61.2 & 61.9 & 45.3 & 39 \\ 
PL++: AVOD \cite{you_Pseudo-LiDAR++_2020} & Depth+Flow & 500ms & 89 & 77.8 & 69.1 & 63.2 & 46.8 & 39.8 \\ 
PL++: P-RCNN \cite{you_Pseudo-LiDAR++_2020}  & Depth+Flow & 510ms & 88 & 73.7 & 67.8 & 62.3 & 44.9 & 41.6 \\ 
OC-Stereo \cite{Pon_ocstereo_2020}& Depth+Instance Mask & 350ms & 89.65 & 80.03 & 70.34 & 64.07 & 48.34 & 40.39 \\ 
ZoomNet \cite{xu_Zoomnet_2020} & Depth+Instance Mask & - & 90.44 & 79.82 & 70.47 & 62.96 & 50.47 & 43.63 \\ 
Disp R-CNN \cite{sun_disprcnn_2020} & Depth+Instance Mask+CAD & 425ms & 90.47 & 79.76 & 69.71 & 64.29 & 47.73 & 40.11 \\

\hline
TL-Net \cite{TL-net}& None & - & 59.51 & 43.71 & 37.99 & 18.15 & 14.26 & 13.72 \\

Stereo RCNN \cite{Li_StereoRCNN_CVPR_2019} & None & 417ms & 85.84 & 66.28 & 57.24 & 54.11 & 36.69 & 31.07 \\ 
IDA3D \cite{Peng_IDA3D_CVPR_2020}&None&300ms&87.08 &74.57 &60.01 &54.97 &37.45 &32.23\\
RTS3D(iteration=2,resl =10) \cite{Li_RTS3D_AAAI_2021} & None & 22ms & 90.26 & 77.23 & 68.28 & 63.65 & 44.5 & 37.48 \\ 
Ours(iteration=2, resl =10) & None & 28ms &  \textbf{90.45(+0.19)} & \textbf{79.36(+2.13)} &\textbf{70.34(+2.06)} & \textbf{65.26(+1.61)}& \textbf{47.07(+2.57)} & \textbf{39.62(+2.14)} \\ \hline
\end{tabular}
}
\label{tab:ap3d}
\end{center}
\end{table*}
\footnotetext{\url{https://github.com/Banconxuan/RTS3D}\label{web}}

\begin{table*}[t]\scriptsize
\begin{center}
\caption{Comparison ($AP_{BEV}$) of state-of-the-art 3D detection methods for car category on KITTI validation set. The number in the bracket indicate the improvement compared to RTS3D\textsuperscript{\ref {web}}
}
\resizebox{\linewidth}{!}{
\begin{tabular}{l|c|c|c|c|c|c|c|c}
\hline
Method & Extra supervision & Time &  \multicolumn{3}{c|}{IoU $ >$ 0.5} & \multicolumn{3}{c}{IoU $>$ 0.7}\\ \cline{4-9}
 &  &  & Easy & Moderate & Hard & Easy & Moderate & Hard \\ \hline\hline
3DOP \cite{3DOP} & Instance Mask &-&55&41.3&		34.6&12.6&9.5&7.6
\\ 
MLF \cite{MLF}& Depth &-&-&53.7&	-&-&	19.5&-
\\ 
PL: F-PointNet \cite{wang_PseudoLiDAR_2019} & Depth+Flow & 670ms &89.8	&77.6&68.2	&72.8&	51.8&44	
\\ 
PL: AVOD \cite{wang_PseudoLiDAR_2019} & Depth+Flow & 510ms &76.8&65.1&56.6	&60.7&39.2&37
\\ 
PL++: AVOD \cite{you_Pseudo-LiDAR++_2020}& Depth+Flow &510ms&89&	77.5&68.7&74.9&56.8	&49
\\ 
PL++: PIXOR \cite{you_Pseudo-LiDAR++_2020}& Depth+Flow & 510ms &89.9&75.2&		67.3& 79.7	&	61.1	&	54.5	
\\   
PL++: P-RCNN \cite{you_Pseudo-LiDAR++_2020} & Depth+Flow & 510ms &88.4&76.6&69	&73.4&	56&52.7	
\\ 
OC-Stereo \cite{Pon_ocstereo_2020} & Depth+Instance Mask&350ms& 90.01&80.63&71.06	&77.66	&	65.95&	51.20	
\\ 
ZoomNet \cite{xu_Zoomnet_2020} & Depth+Instance Mask &-&90.62&88.40		&71.44&78.68&66.19&57.60	
\\ 
Disp R-CNN \cite{sun_disprcnn_2020} & Depth+Instance Mask+CAD & 425ms &90.67&80.45&71.03&77.63&	64.38&	50.68 \\\hline
TL-Net \cite{TL-net} & None&- &62.46&45.99&41.92&29.22	&	21.88&18.83 \\ 
Stereo RCNN \cite{Li_StereoRCNN_CVPR_2019}& None & 417ms & 87.13 &74.11&58.93	&68.50	&48.30	&	41.47	\\ 
IDA3D \cite{Peng_IDA3D_CVPR_2020}&None&300ms&88.05&76.69& 67.29& 70.68& 50.21 &42.93\\
RTS3D(iteration=2,resl =10) \cite{Li_RTS3D_AAAI_2021}& None & 22ms & 90.41& 78.70&70.03& 76.56& 56.46& 48.20\\ 
Ours(iteration=2, resl =10) & None & 28ms & \textbf{90.61(+0.20)}&\textbf{80.50(+1.8)}& \textbf{70.34(+0.31)}&\textbf{77.48(+0.92)}&\textbf{58.41(+1.95)}&\textbf{49.95(+1.75)}
\\ \hline
\end{tabular}
}
\label{tab:apbird}
\end{center}
\end{table*}

\subsection{Non-uniform FCE space generation}
For the stereo images, we adopt the efficient model ResNet-18 \cite{He_ResNet_CVPR_2016} to extract multi-scale feature maps, including low-level texture feature map $F1$, middle-level semantic feature map $F2$, and high-level semantic feature map $F3$. Based on the multi-scale feature maps, we propose high-level semantic enhanced feature consistency embedding (FCE) module with high-level semantic radial basis function (RBF) to exploit more contextual information as follows.

Assuming that $p_i$ is the world coordinate of a point in shape prior non-uniform grid latent space, we convert it into image space as follows:
\begin{equation}
 x_{ij}^{lr} = h_{j} K_{lr}
 \begin{pmatrix} R^{lr} & t^{lr} \\ 0 & 1\end{pmatrix} p_i, j=1,2,3,
\end{equation}

\begin{figure}
\begin{center}
\includegraphics[width=0.7\linewidth]{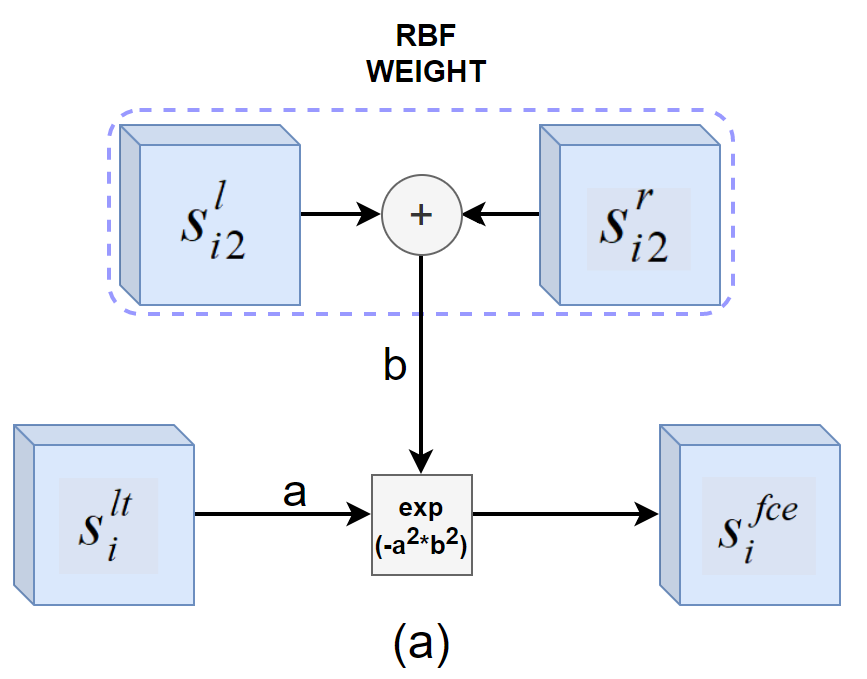}
\includegraphics[width=0.7\linewidth]{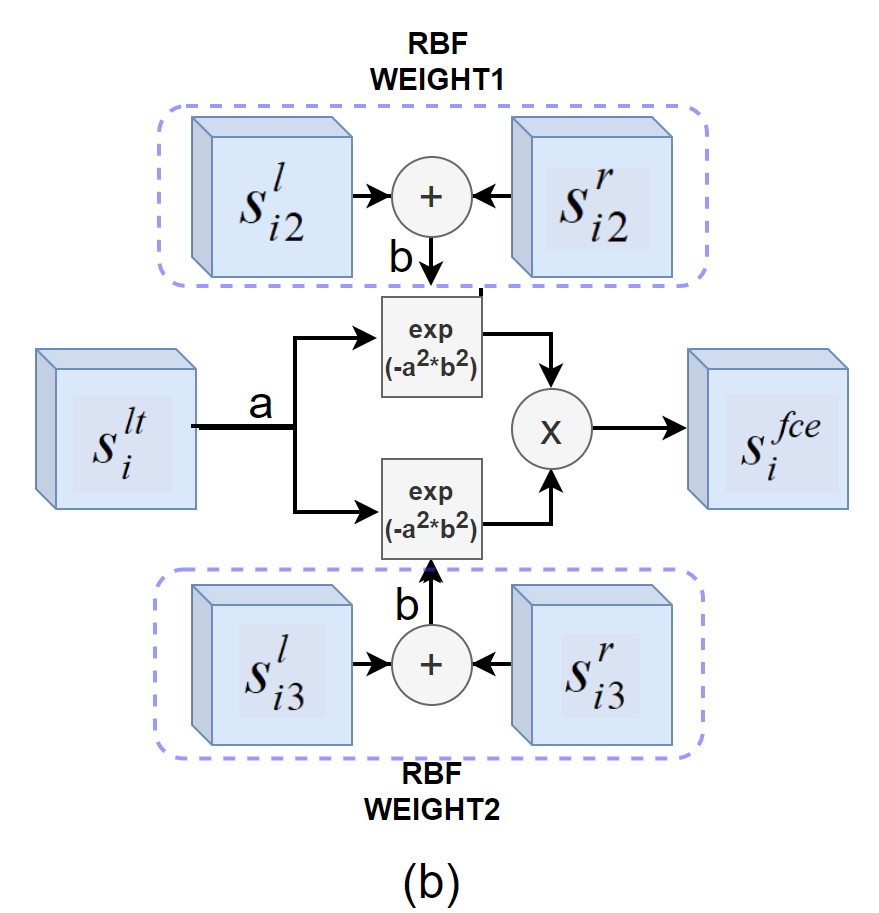}
 \caption{Feature consistency embedding (FCE) module (top) and our high-level semantic enhanced FCE module (bottom). Compared to FCE module, our high-level semantic enhanced FCE module exploits the high-semantic features to extract more contextual information and better suppress the noise.}
\label{fig:multi-RBF}
\end{center}
\end{figure}
where $K$ are camera intrinsic parameters, $R$, $t$ are camera extrinsic parameters (\textit{i.e.,} rotation matrix and translation matrix), $h$ is affine transformation matrix, and $lr$ means the left or right images. Based on the coordinate in image space, we extract multi-scale features $S_{ij}^l$ and $S_{ij}^r$ from both left and right images for point $p_i$.   
\begin{equation}
S_{ij}^l = F_j(x_{ij}^l), \\
S_{ij}^r = F_j(x_{ij}^r), j=1,2,3,
\end{equation}
where $j$ represents the level of the feature map. With the extracted multi-scale features $S_{ij}^l$ and $S_{ij}^r$, we calculate the low-level texture feature $S_{i}^{lt}$, middle-level semantic feature $S_{i}^{ms}$, and high-level semantic feature $S_{i}^{hs}$ for point $p_i$ as follows.
\begin{equation}
\begin{array}{c}
S_{i}^{lt} = S_{i1}^l - S_{i1}^r,     \\
S_{i}^{ms} = (S_{i2}^l + S_{i2}^r)/2,     \\
S_{i}^{hs} = (S_{i3}^l + S_{i3}^r)/2.     \\
\end{array}
\end{equation}
Then, we use two different RBFs to encode the texture and semantic information together to suppress noise.
\begin{equation}
\begin{array}{c}
S_{i}^{fce1} = exp(-(S_{i}^{lt})^2 * (S_{i}^{ms})^2 ),     \\
S_{i}^{fce2} = exp(-(S_{i}^{lt})^2 * (S_{i}^{hs})^2 ).     \\
\end{array}
\end{equation}
Finally, we generate the final enhanced FCE feature as the product of $S_{i}^{fce1}$ and $S_{i}^{fce2}$.
\begin{equation}
\begin{array}{c}
S_{i}^{fce} = S_{i}^{fce1} * S_{i}^{fce2}.
\end{array}
\end{equation}
In shape prior latent space, there are 1000 sampling points. For each point, we caculate the corresponding feature using our high-level semantic enhanced FCE module. As a result, we generate 4D non-uniform FCE space for each proposal.

Fig. \ref{fig:multi-RBF} further compares feature-consistency embedding (FCE) module and  our high-level enhanced FCE module. Compared to original FCE module, we use not only the middle-level semantic cue $F_{2}$ but also the high-level semantic cue $F_{3}$. Thus, more contextual information can be exploited to suppress the noise that is incorporated from the low-level texture cue.

\subsection{3D detection}
As mentioned above, each 3D proposal is represented by the 4D feature in FCE space. Next, we use a 3D object detector to perform 3D detection. To balance speed and accuracy, we employed an improved PointNet \cite{Qi_PointNet_CVPR_2017} as the 3D detector, similar to RTS3D. To improve detection quality, we use a cascaded strategy for refinement, where the output bounding boxes of PointNet can be used as the new input 3D proposals for the nex iteration. In this paper, we adopt two iterations for the experiments.

\begin{figure*}
\begin{center}
\includegraphics[width = 1.0\linewidth]{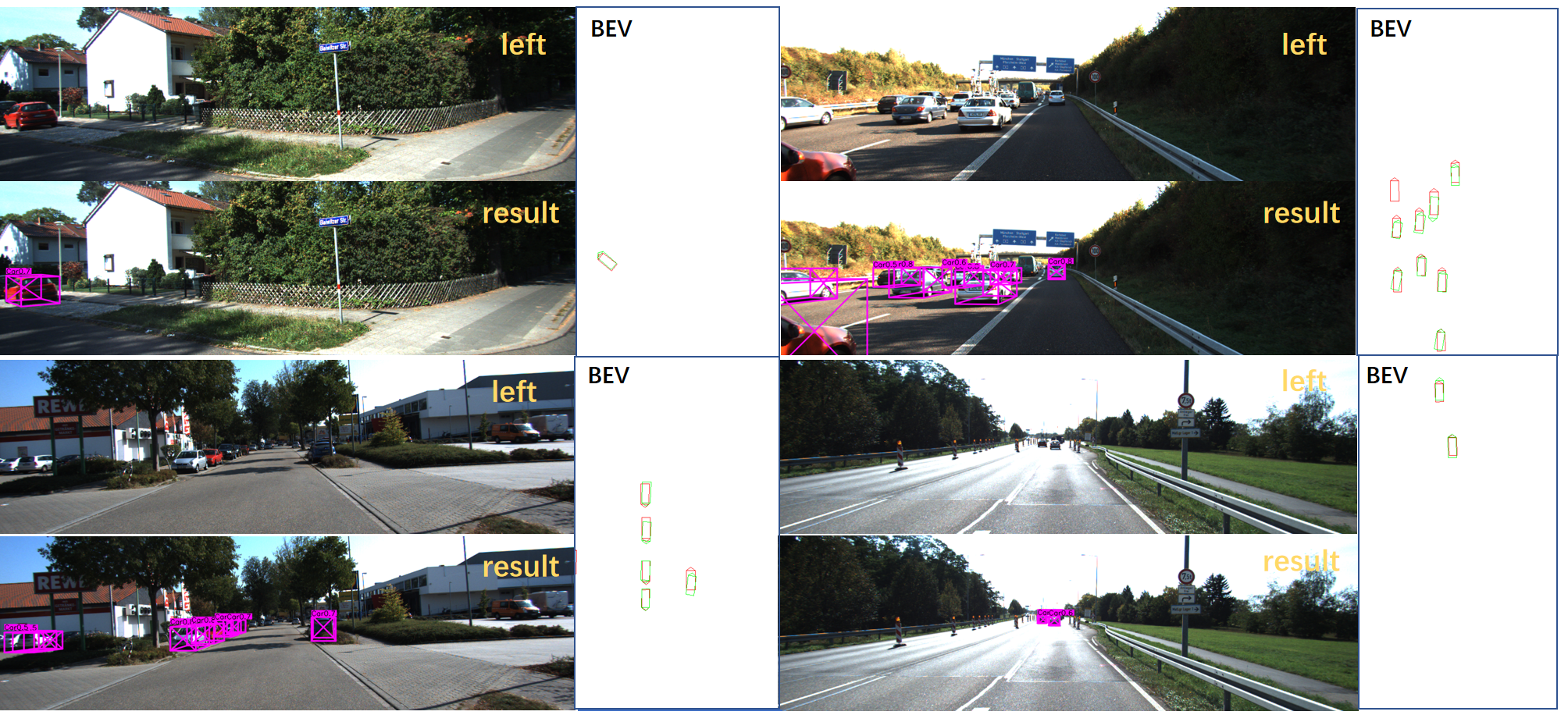}
 \caption{Qualitative results on KITTI validation set. The 3D detection results in left and right images  and the corresponding results in bird view are shown. In the bird view, the red bounding box is GT, and the green bounding box is the detection result.}
\label{fig:results}
\end{center}
\end{figure*}

\section{Experiments}
\subsection {Dataset and implementation details}

In this section, we perform the expriments on the typical KITTI benchmark \cite{kitti} to compare with the state-of-the-art methods and validate the effectiveness of our proposed SPS3D. KITTI benchmark \cite{kitti} is one of the largest computer vision dataset in automatic driving scene. In the task of stereo 3D object detection, it provides stereo images and the corresponding 3D bounding box annotation information. Following the protocol widely used in \cite{Li_StereoRCNN_CVPR_2019,wang_PseudoLiDAR_2019,Pon_ocstereo_2020}, we spilt the original training set into the training set and validation set, respectively. The training set has 3712 images and the validation set has 3769 images.

We adopt the efficient ResNet-18 \cite{He_ResNet_CVPR_2016} as the backbone.  Our method is trained with three NVIDIA TitanX GPUs with Adam for optimization. During the training,  there are 80 epochs and the learning rate is set as 0.000375. To have a fair comparison with RTS3D \cite{Li_RTS3D_AAAI_2021}, we perform the inference on a single NVIDIA 2080Ti GPU. Resl for all experiments is set to 10 and interation for 2.

\subsection{Comparison with state-of-the-art methods}
Here, we compare our SPS3D with some state-of-the-art methods on KITTI validation set, including 3DOP \cite{3DOP},  MLF \cite{MLF}, PL: F-PointNet \cite{wang_PseudoLiDAR_2019}, PL: AVOD \cite{wang_PseudoLiDAR_2019}, PL++: AVOD \cite{you_Pseudo-LiDAR++_2020},  PL++: PIXOR \cite{you_Pseudo-LiDAR++_2020},  PL++: P-RCNN \cite{you_Pseudo-LiDAR++_2020},  OC-Stereo \cite{Pon_ocstereo_2020},  ZoomNet \cite{xu_Zoomnet_2020}, Disp R-CNN \cite{sun_disprcnn_2020}, TL-Net \cite{TL-net}, Stereo RCNN \cite{Li_StereoRCNN_CVPR_2019},  RTS3D \cite{Li_RTS3D_AAAI_2021}, IDA3D \cite{Peng_IDA3D_CVPR_2020},
DSGN \cite{sun_disprcnn_2020}, YOLOstereo3d \cite{liu2021yolostereo3d}.
According to the degree of occlusion and truncation, the validation set is divided into three subsets: \texttt{easy}, \texttt{moderate} and \texttt{hard}.  Table \ref{tab:ap3d} shows the comparison in terms of both speed and accuracy $AP_{3d}$. Our proposed SPS3D achieves the state-of-the-art accuracy, which outperforms the methods without using extra supervision information. For example, on the \texttt{moderate} subset, Stereo RCNN \cite{Li_StereoRCNN_CVPR_2019} and RTS3D \cite{Li_RTS3D_AAAI_2021} has an $AP_{3d}$ scores of 66.28\% and 77.23\%, while our SPS3D has an $AP_{3d}$ score of 79.23\%. Thus, our SPS3D outperforms  Stereo RCNN and RTS3D by an absolute gain of 3.08\% and 2.13\%. To show the superiority of our SPS3D, Table \ref{tab:ap3d} further provides the comparison in terms of both speed and accuracy $AP_{bev}$. Similarly, our proposed SPS3D  outperforms these methods (\textit{e.g.,} RTS3D) without using extra supervision information.

In addition to the high accuracy, our SPS3D has a fast inference real-time speed. For example,  the inference time of Stereo RCNN \cite{Li_StereoRCNN_CVPR_2019} is 417ms, while that of our SPS3D has an $AP_{3d}$ is 28ms. Namely, our SPS3D is almost 14 times faster than Stereo R-CNN. Compared to RTS3D, our SPS3D has a large improvement on accuracy without adding many computational costs. 
We further  show some qualitative results of 3D object detection in Fig. \ref{fig:results}. Our proposed SPS3D can accurately detect the objects of different scales, even in the crowded scenes.

\begin{table}[t]
\begin{center}
\caption{Comparison of three different sampling strategies on KITTI validation set.}
\begin{tabular}{p{3.8cm}|p{1cm}|p{1.2cm}|p{1cm}}
\hline
\centering{Method} & \multicolumn{3}{c}{IoU $>$ 0.7}\\ \cline{2-4}
  & Easy & Moderate & Hard \\ \hline\hline
Uniform sampling  & 63.65 &44.50&37.48 \\ 
Extreme non-uniform sampling  &64.63&46.45&38.92 \\     
Our non-uniform sampling &65.14 &46.86&39.02 \\ 
\hline
\end{tabular}
\label{tab:samplingstrategy}
\end{center}
\end{table}

\subsection{Ablation Study}
In this subsection, we conduct the ablation study to verify the effectiveness of different modules in our SPS3D.

{\bf Shape prior non-uniform sampling (SPS)} We propose shape prior non-uniform stratey to generate the sampling points and construct the shape prior latent space. To demonstrate the effectiveness of non-uniform sampling strategy,  we compare three different strategies (see Fig. \ref{fig:nonsampling}) in Table \ref{tab:samplingstrategy}. Our non-uniform sampling strategy outperforms both the uniform sampling strategy and the extreme non-uniform sampling strategy. It can be concluded as follows. (1) Compared to the inner region, the outer region is more important for 3D detection. (2) Both the inner region and outer region can provide the useful information for 3D detection.

{\bf High-level semantic enhanced module (HSE)} To suppress the noise and extract more contextual information, we propose high-level semantic enhanced FCE module to build FCE space for each proposal. Table \ref{tab:semantic} compare our high-level semantic enhanced FCE module with the original FCE module. Our high-level semantic enhanced FCE module has a better performance. For example, Our semantic enhanced module has 1.95\% improvements on \texttt{moderate} subset.

\begin{table}[t]
\begin{center}
\caption{Effectiveness of our proposed high-level semantic enhanced FCE module on KITTI validation set.}
\begin{tabular}{p{3.8cm}|p{1cm}|p{1.3cm}|p{1cm}}
\hline
\centering{Method} &  \multicolumn{3}{c}{IoU $>$ 0.7}\\ \cline{2-4}
& Easy & Moderate & Hard \\ \hline\hline
Original FCE module  & 63.65 & 44.50 & 37.48 \\ 
Semantic enhanced FCE module   & 64.46&46.45&38.90 \\     
\hline
\end{tabular}
\label{tab:semantic}
\end{center}
\end{table}

\begin{table}[t]
\begin{center}
\caption{Impact of integrating the proposed two modules to the baseline on KITTI validation set.}
\begin{tabular}{p{1.7cm}|p{1.7cm}|p{1cm}|p{1.3cm}|p{1cm}}
\hline
\multicolumn{2}{c|}{Method} & \multicolumn{3}{c}{IoU $>$ 0.7}\\ \hline 
\centering{HSE} & \centering{SPS} & Easy & Moderate & Hard \\ \hline\hline
    &   & 63.65 & 44.50 & 37.48 \\ 
\centering{\checkmark}& &64.46&46.45&38.90 \\ 
& \centering{\checkmark}  &65.14 &46.86&39.02 \\ 
   \centering{\checkmark}   &  \centering{\checkmark}  & 65.26&47.07& 39.62\\     
\hline
\end{tabular}
\label{tab:integration}
\end{center}
\end{table}

\begin{figure}
\begin{center}
\includegraphics[width = 1.0\linewidth]{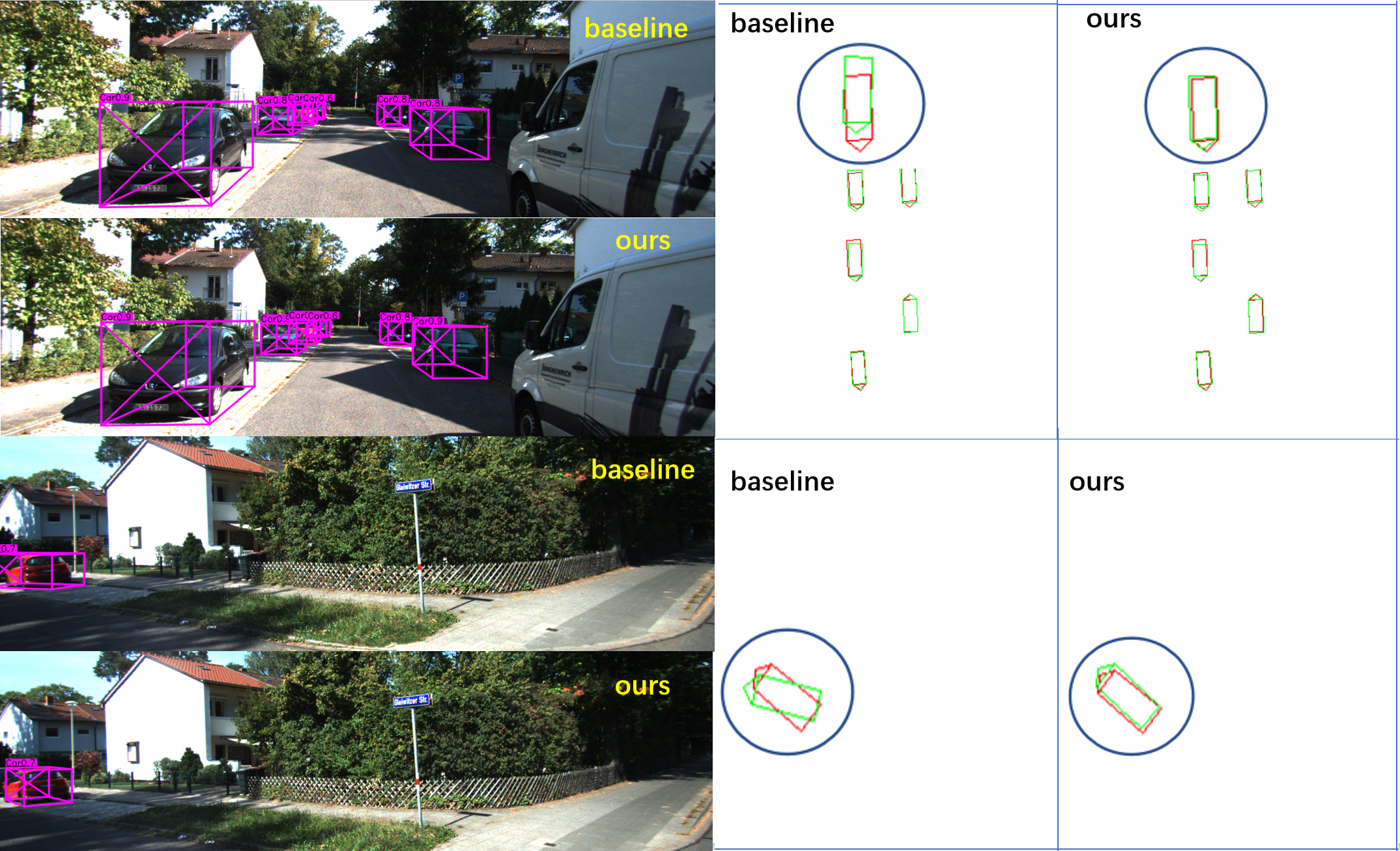}
\caption{Qualitative results of our SPS3D and the baseline. The left column shows 3D detection results on the left image, and the right column shows the detection results in bird’s eye view. In bird’s eye view, the red box represents GT, and the green box represents the detection result. Compared to the baseline, our SPS3D can provide more accurate detection.}
\label{fig:qres}
\end{center}
\end{figure}

{\bf Integration of different modules} Table \ref{tab:integration} shows the impact of integrating the proposed modules, including SPS and HSE, to the baseline RTS3D. Our proposed SPS3D has a large improvement by adding these two modules to the baseline. On the \texttt{moderate} subset, it provides an absolute gain of 2.57\%. Further, we compare the qualitative results of our SPS3D and the baseline in Fig. \ref{fig:qres}.

\section{Conclusion}
In this paper, we have proposed shape prior non-uniform sampling guided stereo 3D object detection. We argue that the outer region is more important for 3D detection. Inspired by this, we propose to perform the non-uniform sampling to generate the latent space and FCE space, where more sampling points are generated from the outer region. In addition, to suppress the noise and exploit more contextual information, we propose high-level semantic enhanced FCE module for consistency feature extraction. Experiments on the KITTI benchmark show that  our proposed method achieves the state-of-art performance at realt-time speed.

\ifCLASSOPTIONcaptionsoff
  \newpage
\fi


\end{document}